\documentclass[conference]{IEEEtran}
\IEEEoverridecommandlockouts
\usepackage{cite}
\usepackage{amsmath,amssymb,amsfonts}
\usepackage{algorithmic}
\usepackage{graphicx}
\usepackage{textcomp}
\usepackage{xcolor}

\usepackage{multirow}
\usepackage{subfigure}
\usepackage{booktabs}
\usepackage[colorlinks,linkcolor=red]{hyperref}
\usepackage[misc]{ifsym} 

\def\BibTeX{{\rm B\kern-.05em{\sc i\kern-.025em b}\kern-.08em
    T\kern-.1667em\lower.7ex\hbox{E}\kern-.125emX}}
\begin{document}

\title{Multi-Scale Context Aggregation Network with Attention-Guided for Crowd Counting\\
}

\author{
\IEEEauthorblockN{Xin Wang\textsuperscript{1,2}, Rongrong Lv\textsuperscript{1}, Yang Zhao\textsuperscript{2}, Tangwen Yang\textsuperscript{1}$^{\textrm{\Letter}}$, Qiuqi Ruan\textsuperscript{1,2}}
\IEEEauthorblockA{\textsuperscript{1}\textit{Institute of Information Science, School of Computer and Information Technology} \\
\textit{Beijing Jiaotong University,} Beijing 100044, China \\
\textsuperscript{2}\textit{Guangdong Key Laboratory of Intelligent Information processing}\\
\textit{Shenzhen University,} Shenzhen, Guangdong 518060, China \\
E-mail: \{19112007, 18120309, twyang, qqruan\}@bjtu.edu.cn, zhaoyangmaths@163.com}
}

\maketitle

\begin{abstract}
    Crowd counting aims to predict the number of people and generate the density map in the image. 
    There are many challenges, including varying head scales, the diversity of crowd distribution across images and cluttered backgrounds. In this paper, we propose a multi-scale context aggregation network (MSCANet) based on single column encoder-decoder architecture for crowd counting, which consists of an encoder based on dense context-aware module (DCAM) and a hierarchical attention-guided decoder. To handle the issue of scale variation, we construct the DCAM to aggregate multi-scale contextual information through densely connecting the dilated convolution with varying receptive fields. The proposed DCAM can capture rich contextual information of crowd areas due to its long-range receptive fields and dense scale sampling. Moreover, to suppress the background noise and generate a high-quality density map, we adopt a hierarchical attention-guided mechanism in the decoder. This helps to integrate more useful spatial information from shallow feature maps of the encoder by introducing multiple supervision based on semantic attention module (SAM). Extensive experiments demonstrate that the proposed approach achieves better performance than other similar state-of-the-art methods on three challenging benchmark datasets for crowd counting. The code is available at \url{https://github.com/KingMV/MSCANet}.
\end{abstract}


\section{Introduction}
\begin{figure}[t]
	\centering
	\subfigure{
		\includegraphics[width=0.95\linewidth]{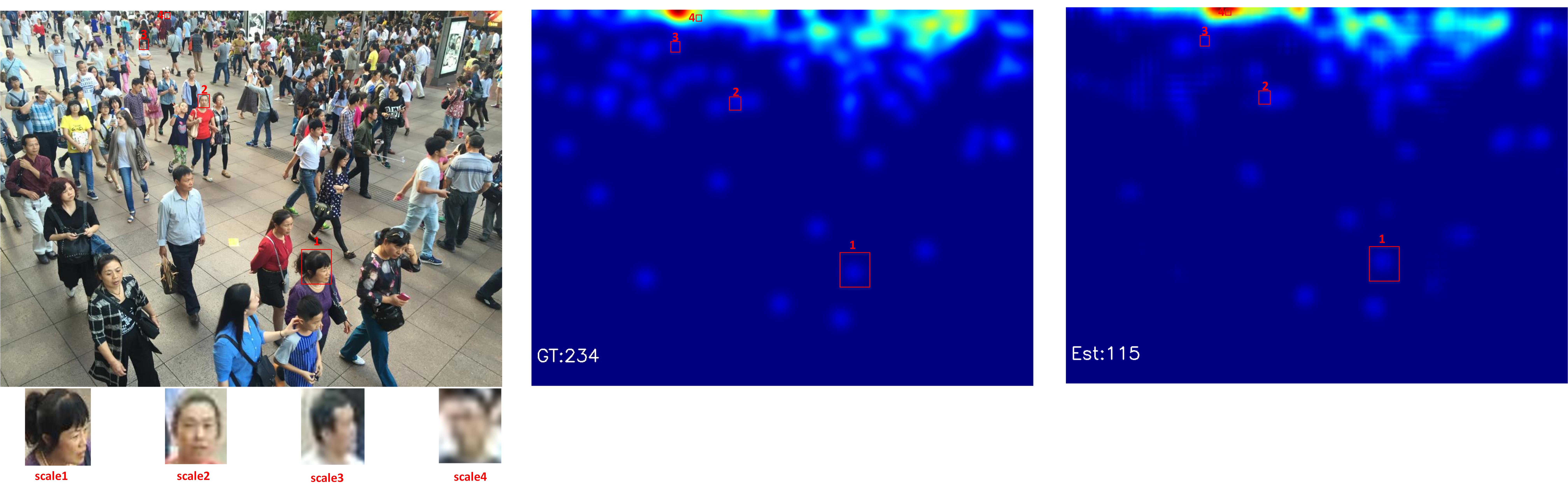}
    }
	\subfigure{
		\includegraphics[width=0.95\linewidth]{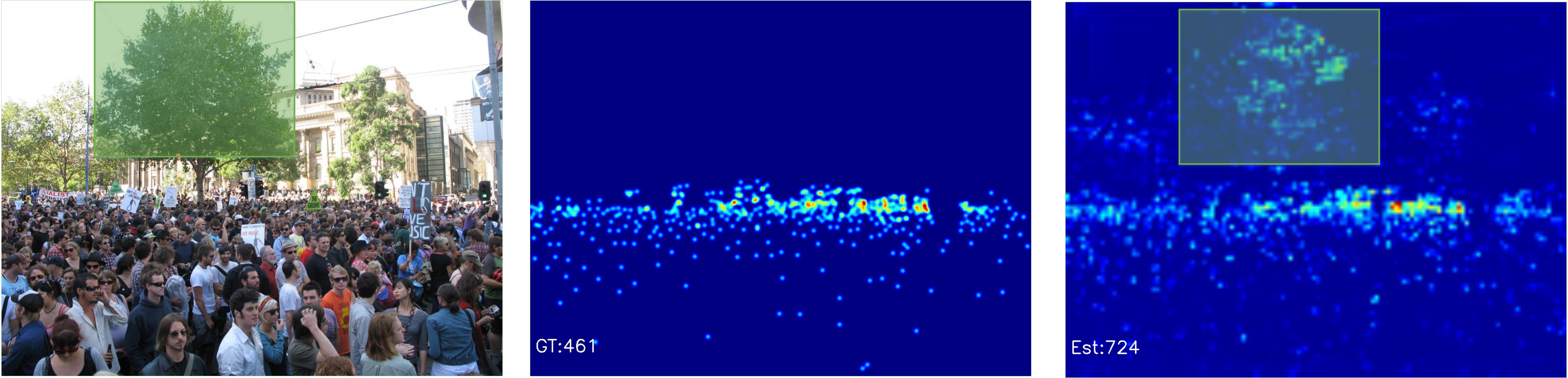}
    }
    \caption{Some samples contain scale variations and background clutter in the ShanghaiTech dataset. The red rectangle indicates some human heads of different sizes. 
    The green rectangle contains some clutters similar to the crowd area, especially the high-density crowd. The first column shows the original image, the second column shows the ground-truth density map, and the third column is the predicted density map from CSRNet method\cite{li2018csrnet} in 2018.}
    \label{fig1}
\end{figure}

Crowd counting has attracted much attention in recent years due to its important application including video surveillance, public security, et al. 
In addition, it is a key technique for high-level behavior analysis algorithms, such as crowd behavior analysis, crowd gathering detection. Specifically, crowd density estimation is also beneficial to prevent the spread of the 2019-nCoV virus. 
However, scale variations, huge crowd diversity, and background clutter are critical challenges of crowd counting. As shown in Figure \ref{fig1}, the head scale within an image is varying due to camera perspective, and the crowd distribution across scenes also represents different patterns. Some CNN-based methods usually overestimate the density map of backgrounds due to the complexity of backgrounds, as analyzed in some crowd counting review papers\cite{gao2020cnn,sindagi2018survey}. Besides, some gridded areas (such as trees and buildings) are more likely to be mistaken in density map because the appearance of backgrounds is very similar to that congested crowd areas.
To address the scale variations issue, many multi-column network based methods\cite{zhang2016single, sam2017switching, guo2019dadnet,jiang2019crowd} are proposed to extract multi-scale features, where different column networks are designed with different kernel sizes.
However, these multi-column based methods have a more bloated structure, which lead to redundant information of each subnetwork, as analyzed in CSRNet \cite{li2018csrnet}. Besides, inspired by inception architecture, some scale-aware modules\cite{cao2018scale, chen2019scale} adopt multiple various convolution kernels with different receptive fields to extract features at various scales. These modules can be plugged directly into the existing single column network. 
The advantage of single-column-based methods is their elegant network structure and high training efficiency.
However, the rate of dilated kernel at different columns needs to be carefully selected in these scale-aware modules, which is challenging to capture various continuous scales.
We observe that a small object of crowded areas can be represented by its neighbor information and the rich contextual information at multiple scales can improve counting accuracy. Therefore, in this paper, we investigate how to extract multi-scale contextual information from different receptive fields.

The attention mechanism is usually used to make the foreground area get more attention in \cite{liu2019adcrowdnet,miao2020shallow,liu2018decidenet,gao2019pcc,jiang2019learning}. The common idea is to design a single attention model with a bloated structure to predict an attention map and then multiply the estimated density map by the predicted attention map. However, this architecture makes the whole model too complicated and leads to a heavy computing burden. We observe that the shallow feature maps contain many edge and background noise  information. Therefore, it is necessary to suppress these noise information in the shallow feature maps.

In this paper, we propose a multi-scale context aggregation network (MSCANet) based on encoder-decoder architecture for crowd counting, which can handle the scale variations and generate a high-quality density map. 
Specifically, we design a dense context-aware module (DCAM) to aggregate multi-scale contextual information by densely connecting the dilated kernels of different receptive fields. 
Compared with existing scale-aware modules that only extract limited scale context information, the proposed DCAM can capture rich contextual information and scale diversity due to its larger and finer range of receptive fields. 
Moreover, inspired by UNet\cite{ronneberger2015u} and FPN\cite{lin2017feature}, we present a hierarchical attention-guided decoder to hierarchically integrate different feature maps of the encoder for a high-quality density map. 
The multiple lightweight semantic attention modules (SAM) are added into different fusion layers of the decoder, which guide the model to pay more attention to the shallow feature maps of the crowd area. The contributions of this paper are summarized as follows:
\begin{itemize}
\item We propose a multi-scale context aggregation network based on encoder-decoder for crowd counting, which can improve the
multi-scale representation and generate a high-quality density map. 
\item Multi-scale context aggregation module is designed to help the encoder capture rich context and broad range scale information. Specifically, the module consist of multiple dense context-aware modules (DCAM) stacked, which densely connects multiple dilated kernels with different receptive fields.
\item We propose a hierarchical attention-guided decoder to explicitly integrate important information of different feature maps from the encoder for a high-quality density map. Multiple SAMs are introduced into different stages of the decoder to make the shallow feature maps of crowd areas get more attention.
\item Extensive experiments are conducted on three crowd datasets. The results demonstrate that our proposed method achieves better performance than other similar state-of-the-art methods.
\end{itemize}

\section{Related Works}
\subsection{Multi-scale feature extraction}\label{msfe section}
To address the issues of scale variations, researchers proposed many multi-column based methods \cite{zhang2016single}\cite{sam2017switching}to explicitly utilize different columns with different respective fields to extract multi-scale features. Zhang et al.\cite{zhang2016single} first adopted multi-column convolutional neural network (MCNN) to handle scale variations, where the multi-scale features are extracted by three branches with kernels of different sizes (large, medium, small). Sam et al\cite{sam2017switching} proposed an improved method switch-CNN based on MCNN, which introduces a switch classifier to choose an optimal branch for input image patch. However, it is difficult to train multi-column networks and the efficiency of each network is low. Therefore, the single-column deeper network architecture is usually used in many state-of-the-art methods. Among these single-column methods\cite{li2018csrnet}\cite{chen2019scale}\cite{guo2019dadnet}, dilated convolution kernel and deformable convolution kernel are used to build multi-scale aware module. 
Li et al.\cite{li2018csrnet} utilized multiple dilated convolutional layers in the single-column network to increase respective fields without the loss of spatial information. 
Chen et al.\cite{chen2019scale} proposed a Scale Pyramid Module(SPM) to extract multi-scale features, which employs dilated convolution with different rates in parallel. Guo et al.\cite{guo2019dadnet} explored a scale-aware attention fusion with various dilated rates to capture different visual information of crowd regions of interest, and utilized deformable convolutions to generate a high-quality density map.
However, in the above methods, since the rate of dilated kernel at different columns is fixed, the existing scale-aware methods only capture contextual information of specific receptive fields, which limits the representation of more scales.

\subsection{Attention mechanism}
The attention mechanism's goal is to make the model focus on useful information to improve the counting performance. Liu et al.\cite{liu2018decidenet} proposed a DecideNet to combine the estimated density map with detection map through an attention guide. Liu et al.\cite{liu2019adcrowdnet} designed an attention map generator to provide the mask region of the crowd area, and then a multi-scale deformable network is used to predict the density map of crowd mask area. Miao et al.\cite{miao2020shallow} built a shallow features based dense attention network to diminish the impact of the background noise. Chen et al.\cite{chen2020crowd} proposed a crowd attention convolutional network (CAT-CNN) for crowd counting, where the human in the estimated density map can get more attention by encoding a confidence map. However, these existing methods design a single attention model with a complicated structure to enhance the attention of foreground areas.

\begin{figure*}[ht]
	\centering
    \centerline{\includegraphics[width=1.0\linewidth]{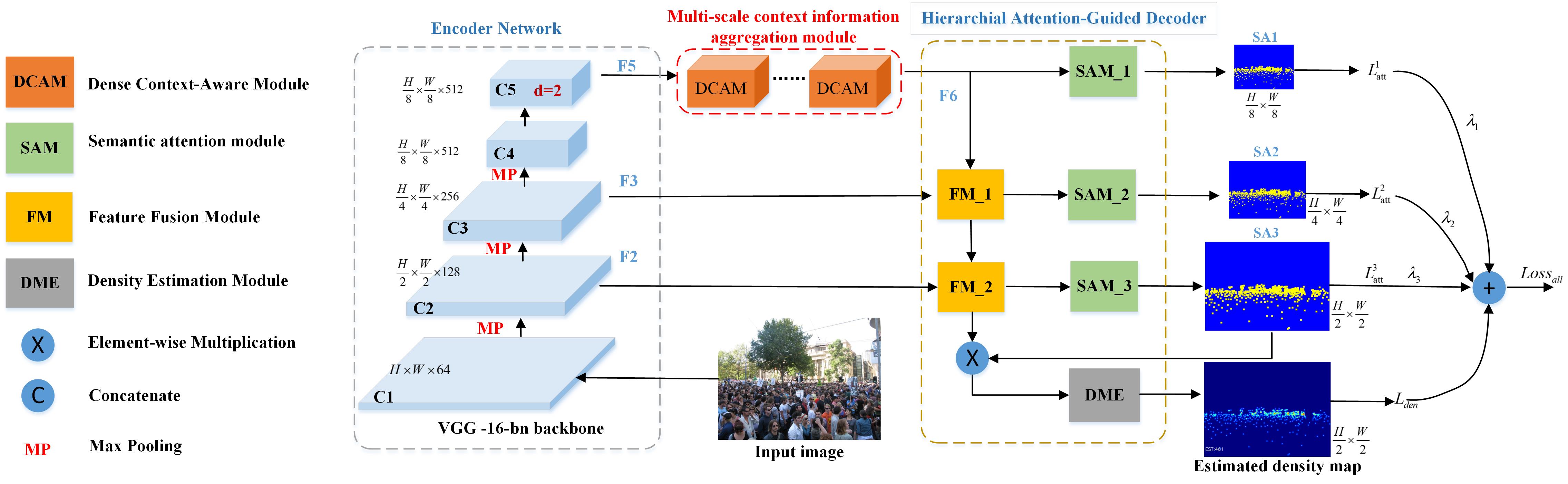}}
    \caption{Overview of the proposed crowd counting framework. The backbone based on VGG16 is used to extract the basic visual feature representation of input image. Then these features are fed into multiple dense context-aware modules (DCAM) to enhance the representation of multi-scale features. Each DCAM aggregates contextual information from different receptive fields through densely connecting dilated convolution of different dilated rates. The hierarchical attention-guided decoder is used to explicitly integrate important information of different feature maps from the encoder.}
    \label{architecture}
\end{figure*}

\section{Proposed Method}
The proposed method is illustrated in Figure \ref{architecture}. The key components are multi-scale context aggregation module and hierarchical attention-guided decoder.
\begin{figure}[t]
	\centering
    \centerline{\includegraphics[width=\linewidth]{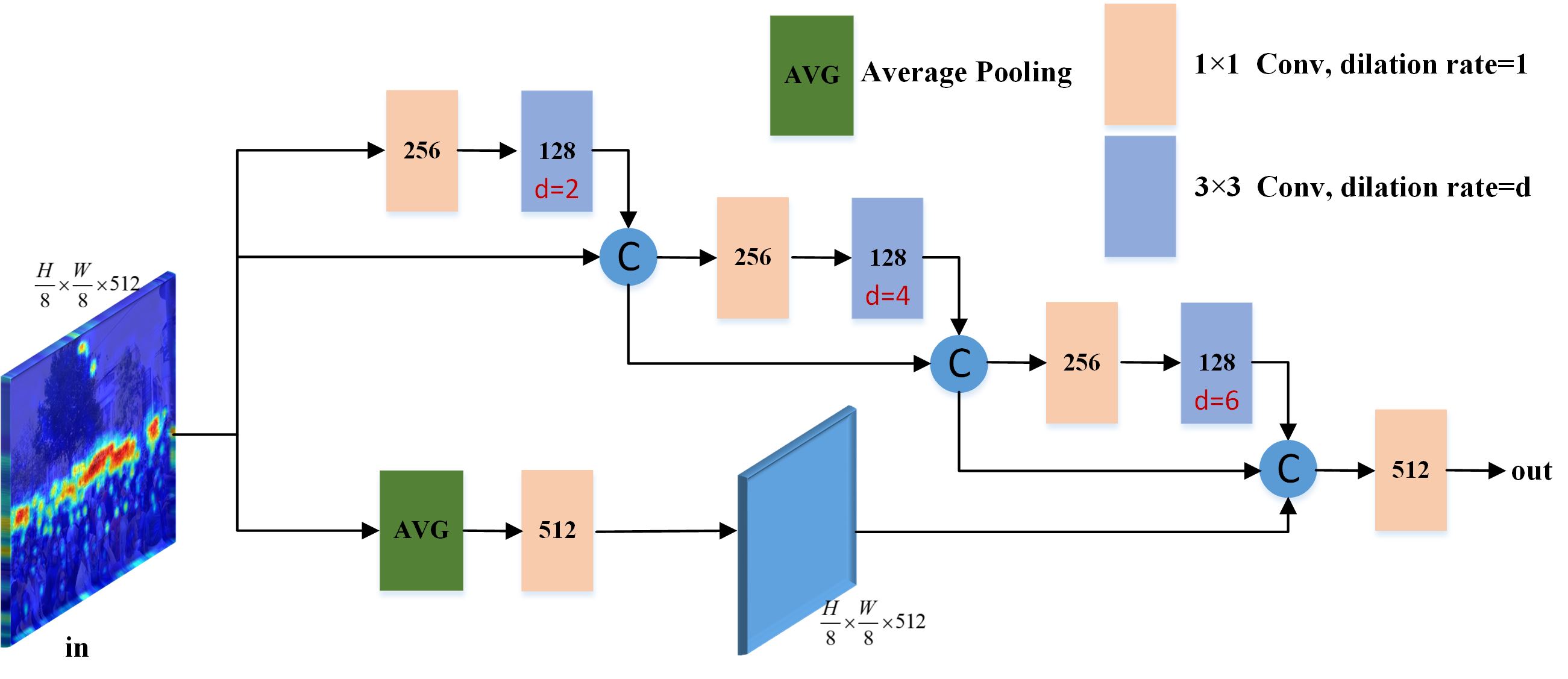}}
    \caption{Illustration of dense context-aware modules (DCAM).}
    \label{DCAM}
\end{figure}

\subsection{Multi-scale Feature Encoder Network (MSFEN)}

Scale pyramid module based on the dilated convolution kernel is a common technology to extract multi-scale features. 
Since the receptive field of each column is fixed in advance, these methods only extract limited different scales.
Inspired by \cite{guo2019dadnet}, we observe that contextual information is very important for high-density crowd area, particular for the pixel-regression task. Therefore, we design a multi-scale context aggregation module to aggregate multi-scale contextual information of crowd through the dilated convolution. The encoder contains two parts: feature extraction backbone and multi-scale context aggregation module that consists of multiple dense context dense-aware modules.

\subsubsection{Feature extraction backbone}
Similar to previous crowd counting works\cite{li2018csrnet}\cite{liu2019adcrowdnet}\cite{chen2019scale}, we adopt the first ten layers of VGG-16 model\cite{simonyan2014very} pre-trained from the ImageNet dataset as the backbone for their strong transfer learning ability. The last two pooling layers and all full connection layers of VGG-16 are removed. The backbone consists of 5 blocks $\{C1,C2,C3,C4,C5\}$. 
The all dilated rates in $C5$ block are set to 2, which can enlarge the receptive fields and reduce the loss of feature map information.

\subsubsection{Dense context-aware module}
The dense context-aware module (DCAM) densely connects multiple dilated convolutions with different dilated rates to integrate multiple contextual information of various receptive fields.
It is well known that the dilated convolution with rate $d$ is able to enlarge the receptive field of $k\times k$ kernel to $k+(k-1)\times(r-1)$ without the reduction of feature map sizes. In \cite{chen2019scale}, the dilated rates in four column branches are usually set to $2,4,6,8$ or other fixed sizes. However, contextual diversity is restricted by the number of branches. Too large receptive field size and too sparse context sampling can degrade model performance. To capture detailed contextual information, we stack multiple dilated convolution layers by dense connection to enlarge the range of receptive fields.

As shown in Figure \ref{DCAM}, the proposed DCAM consists of three dilated convolution layers with dense connection. Considering that the head scale variations are continuous, the dilated rates of DCAM are $2,4,6$, receptively.
The output of $l$-th dilated convolution layer $H_{l}$ is concatenated with all outputs from preceding layers$\{H_1,H_2,\cdots,H_{l-1}\}$, and then they are fed into the next dilated convolution layer. Besides, we employ the global average pooling layer $F_{avg}$ to capture global contextual information of input $X$, and then concatenate with the outputs of different dilated layers. Finally, a $1\times 1$ convolution is used to fuse the concatenated feature maps. Moreover, in order to process more scale diversity, we stack three DCAMs to further enhance the fusion of different contextual information. The contextual information $Y$ can be obtained by the following:
\begin{align}
        H_l &= F_{k,d_l}^{(l)}([x,H_1,H_2,\cdots,H_{l-1}];\Phi) \\
        Y &= F_{1,1}(U_b(F_{avg}(X))+Concat(H_1,H_2,\cdots,H_l))
\end{align}
where $F_{k,d}(\cdot)$ denotes a  $k\times k$ dilated convolution with dilated rate $d$.
$\Phi$ is the parameter of each convolution kernel. $U_b$ is a bilinear upsampling operation. 
\begin{figure}[t]
	\centering
    \centerline{\includegraphics[width=0.95\linewidth]{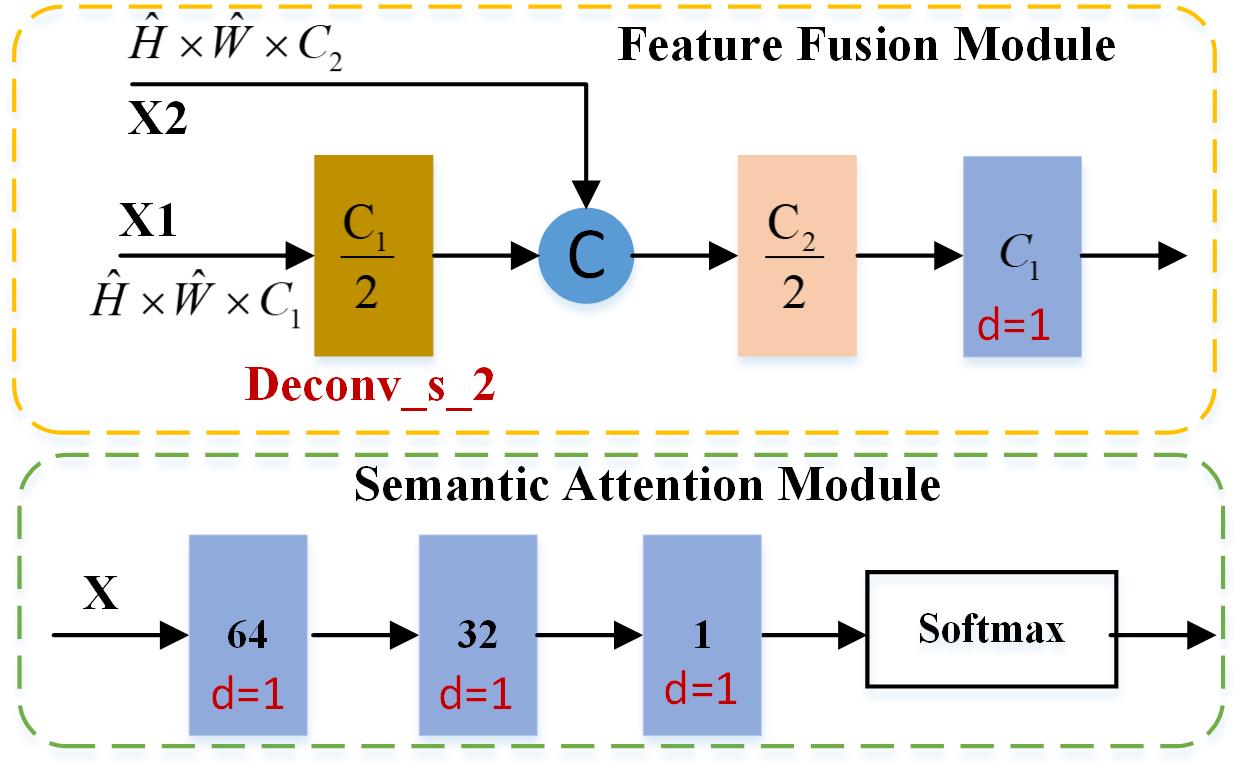}}
    \caption{Illustration of semantic attention module and feature fusion module of Decoder.}
    \label{SAM} 
\end{figure}



\subsection{Hierarchical Attention-Guided Decoder}
The hierarchical attention-guided decoder (HAGD) is proposed to progressively integrate different feature maps of the encoder for a high-quality density map. 
Generally, features of small-scale people can be extracted at an earlier layer. But the earlier layer contains a lot of noises caused by some backgrounds (trees, buildings). Therefore, it is necessary to suppress these background noises in the process of feature fusion. Based on this motivation, we adopt an attention-guided mechanism to increase the attention of crowd area in shallow feature maps through a lightweight semantic attention module (SAM). Similar with UNet structure\cite{ronneberger2015u}, we integrate feature maps $F6$ and $F3$ through a feature fusion module (FM), and then features $M_{3,6}$ from $FM\_1$ are passed into next layers to concatenate with feature map $F2$ through $FM\_2$.
As shown in Figure \ref{SAM}, the structure of SAM is: $CBR(64,3)-CBR(32,3)-CB(1,1)-Sigmoid$. Where $CBR(m,n)$ denotes each convolution layer with $m$ filters of size $n\times n$, followed by Batch Normalization layer and Relu layer. In the last layer, the probability of each pixel belongs the foreground is obtained by a sigmoid function. 


To suppress background noise of the final density map, the element-wise multiplication is conducted to fuse feature map $M_{2,3}$ from $FM\_2$ and estimated attention map from SAM. The final density map is obtained through a density estimation module (DME). 
The DME consists of $3\times 3\times 64$, $3\times 3\times 32$ and $1\times 1\times1$ convolution, each convolution layer is followed by BN layer and Relu Layer.
\begin{equation}
    Y^{p}=DME(M_{2,3}\odot D^{p};\Phi)   \label{ypred}
\end{equation}
where $Y^{p}$ is the estimated density map. $D^{p}$ is the predicted foreground mask from SAM. $\odot$ represents the element-wise multiply operation.

From a hierarchical ensemble perspective, we take each fused feature of decoder into SAM to give foreground more attention. In the process of training, multiple supervision losses of SAM at different scales are calculated to improve the location accuracy and reduce the difficulty of model learning. The ground-truth attention mask $D^{gt}$ is obtained from ground-truth density map $Y^{gt}$. It can be represented as:
\begin{equation}
    D^{gt}(x_i)=
    \begin{cases}
    1& t\geq Y^{gt}(x_i)\\
    0& t\leq Y^{gt}(x_i)
    \end{cases}
\end{equation}
where $D^{gt}(x_i)$ is the ground-truth value at pixel $x_i$ and it is a binary map. $t$ is a threshold value of 0.0001. 
To obtain ground-truth attention mask of different scales, we resize the $D^{gt}$ to $1/2$, $1/4$ and $1/8$ of its original size.

\subsection{Loss function}
The SAM aims to predict the probability of each pixel belongs foreground, it can be seen as a semantic segmentation with two label. Therefore, we train the SAM with a binary cross-entropy loss. It is defined as:
\begin{equation}
    L_{att}=\frac{1}{N}\sum_{i}^{N}-D^{gt}_{i}\log(D^{p}_i)-(1-D^{gt}_i)\log(1-D^{p}_{i}) 
\end{equation}
where $N$ is the number of training samples. $D^{p}_i$, $D^{gt}_i$ are the predicted mask and ground-truth mask of input sample $i$, receptively.

We combine multiple losses of SAM at different scales to give more attention to crowd areas in the feature fusion. Besides, the Euclidean distance between predicted density map $D_i$ and ground-truth density map $Y_i^{GT}$ is used to define the regression loss $L_{den}$. Therefore, the final loss $L$ is the combination of multiple semantic losses and regression loss. 
It can be written as:
\begin{equation}
    L_{den}=\sum_{i=1}^{N}{\left \|D_i-Y_i^{GT}\right \|^2}
\end{equation}
\begin{equation}
    L=\lambda_1L_{att}^1+\lambda_2L_{att}^2+\lambda_3L_{att}^3 \label{L-w}+L_{den}
\end{equation}
where $\lambda_1,\lambda_2,\lambda_3$ are parameter to adjust the weight of SAM at different scales, which are set to $1e^{-2}$, $1e^{-3}$ and $1e^{-4}$ in our experiment. $L_{att}^1$, $L_{att}^2$, $L_{att}^3$ represent the attention loss of SAM at different levels, respectively.

\section{EXPERIMENTS}
\subsection{Ground truth density map}
Following some previous works\cite{li2018csrnet}\cite{cao2018scale}\cite{chen2019scale}, we use a Gaussian kernel to blurring each head annotation to generate the density map of image. For a head at pixel $x_i$, it can be represented in the image by a delta function $\delta(x-x_i)$, and the density map is obtained by convolving with Gaussian kernel$G_{\sigma}(x)$ as follow.
\begin{equation}
    Y_i^{Gt}(x)=\sum_{x_i}\delta(x-x_i)\ast G_{\sigma_i}(x),\ \sigma_i=\beta\bar{d_i}
\end{equation}
where $G_{\sigma_i}(x)$ is a geometry-adaptive Gaussian kernel with variance $\sigma_i$; the variance $\sigma_i$ of each pixel is determined by the average distances $\bar{d_i}$ to its $k$ nearest neighbors. For ShanghaiTech Part A datatset, we use the geometry-adaptive kernel to generate density map. The average distance $\bar{d_i}$ is computed by KNN (K=3) method and the parameter $\beta$ is set to 0.3. For the ShanghaiTech Part B, the UCF\_CC\_50, the UCF-QNRF, $\sigma$  is set to 15.

\begin{table*}[htbp]
    \centering
    \caption{The Comparison results with some state-of-the-art methods on ShanghaiTech, UCF-QNRF and UCF\_CC\_50 datasets. CSRNet* and SPN* denote the reproduction results in our experiment. SHTA and SHTB denote ShanghaiTech Part\_A and Part\_B dataset.}
    \label{SHT}
    \begin{tabular}{lllccccccc}
    \toprule
    \multirow{2}{*}{Method} & \multirow{2}{*}{Year\&Venu} & \multicolumn{2}{c}{SHTA} & \multicolumn{2}{c}{SHTB} & \multicolumn{2}{c}{UCF-QNRF} & \multicolumn{2}{c}{UCF\_CC\_50} \\ \cline{3-10}
    &                     &MAE$\downarrow$  &RMSE$\downarrow$    &MAE$\downarrow$  &RMSE$\downarrow$ &MAE$\downarrow$   &RMSE$\downarrow$     &MAE$\downarrow$    &RMSE$\downarrow$          \\
    \midrule
    MCNN\cite{zhang2016single}    & 2016 CVPR   &110.2 &173.2  &26.4  &41.3  &-     &-      &377.6   &509.1           \\  
    CSRNet\cite{li2018csrnet}  & 2018 CVPR   &68.2  &115.0  &10.6  &16.0  &-     &-  &266.1   &397.5            \\
    SANet\cite{cao2018scale}   & 2018 ECCV  &67.0  &104.5  &8.4   &13.6  &-     &-      &258.4   &334.9             \\
    ADCrowdNet\cite{liu2019adcrowdnet}   & 2019 CVPR   &66.1  &102.1  &7.6   &13.9  &-     &-      &257.9   &357.7             \\
    CANNet\cite{liu2019context}  & 2019 CVPR   &62.3  &100.0  &7.8   &12.2  &107   &183    &212.2   &243.7             \\
    TEDNet\cite{jiang2019crowd}  & 2019 CVPR   &64.2  &109.1  &8.2   &12.8  &113   &188    &249.4   &354.5             \\
    SPN+L2SM\cite{xu2019learn} & 2019 ICCV  &64.2  &98.4   &7.2   &11.1  &104.7 &173.6  &188.4   &315.3             \\
    SPN\cite{chen2019scale}      & 2019 WACV  &61.7  &99.5   &9.4   &14.4   &-     &-      &259.2   &335.9         \\
    SDANet\cite{miao2020shallow}   & 2020 AAAI  &63.6  &101.8  &7.8   &10.2   &-     &-      &227.6   &316.4         \\
    \hline
    CSRNet*  & -           &68.2  &111.6  &10.4     &15.8     &117.1    &216.6   &281.8   &403.3            \\
    SPN*     & -           &63.1  &110.9   &9.6   &14.8   &-     &-      &273.2   &351.9         \\
    MCANet(ours) &-   &60.1  &100.2   &6.8   &11.0   &100.8     &185.9      &181.3   &258.6          \\
    \bottomrule
    \end{tabular}
\end{table*}


\subsection{Implements Details}
Adam optimizer is employed to minimize the loss of model for 500 epochs. The initial learning rate is set to 1e-4 and then is decreased by a factor 0.1 every 10 epochs. 
The first 10 layer's weight of pre-trained VGG16 are used to initialize the trainable parameters of backbone, and other parameters are initialized by a Gaussian distribution with zero mean and a standard deviation of 0.01. All training images are randomly cropped with a fixed size $320\times320$ to train the model. The size of mini-batch is 16.
Besides, the training images and density maps are horizontally flipped with a probability of 0.5. In the test phase, the whole image is sent to the model to predict the density map of the input image.

\subsection{Evaluation metric}
The MAE and RMSE metric are used to evaluate the performance of the proposed model, which measure the accuracy and robustness of model. They are defined as follow:
\begin{equation}
    MAE = \frac{1}{N} \sum_{i=1}^{N}|Y_i^{GT}-Y_i^{pred}|
\end{equation}
\begin{equation}
    RMSE = \sqrt{\frac{1}{N} \sum_{i=1}^{N}|Y_i^{GT}-Y_i^{pred}|^2}
\end{equation}
where $N$ is the number of test image. $Y_i^{GT}$ and  $Y_i^{pred}$ is the ground truth and the estimated counts for $i$-th image,  receptively.

\begin{figure}[t]
	\centering
	\subfigure{
		\includegraphics[width=0.27\linewidth]{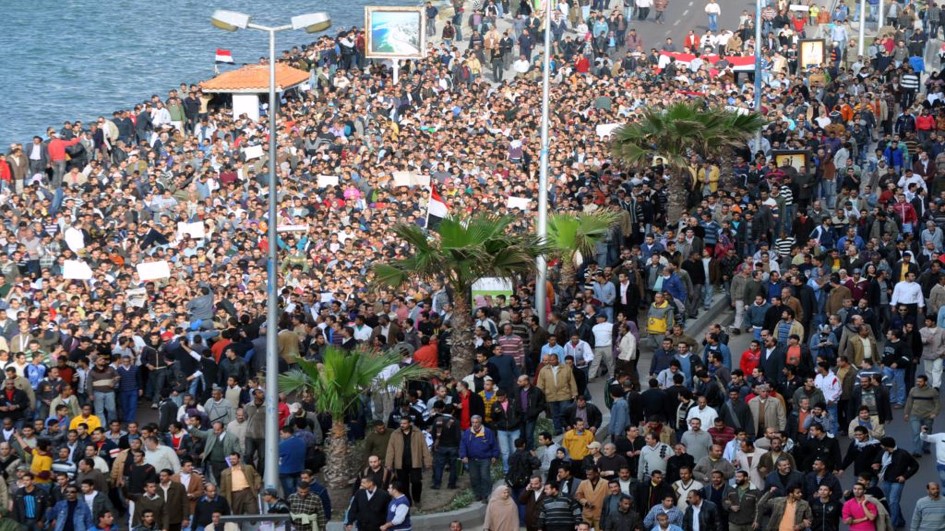}
    }
    \subfigure{
		\includegraphics[width=0.27\linewidth]{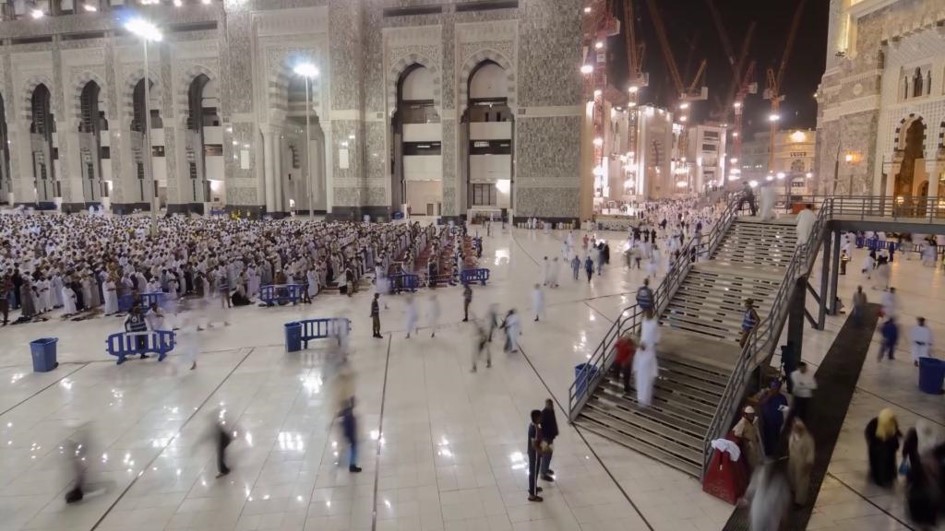}
    }
    \subfigure{
		\includegraphics[width=0.27\linewidth]{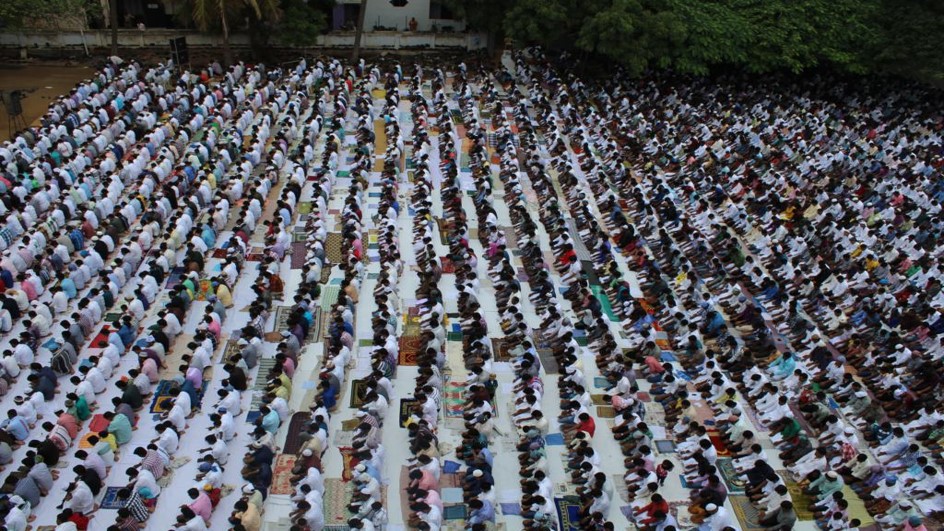}
    }
	\subfigure{
		\includegraphics[width=0.27\linewidth]{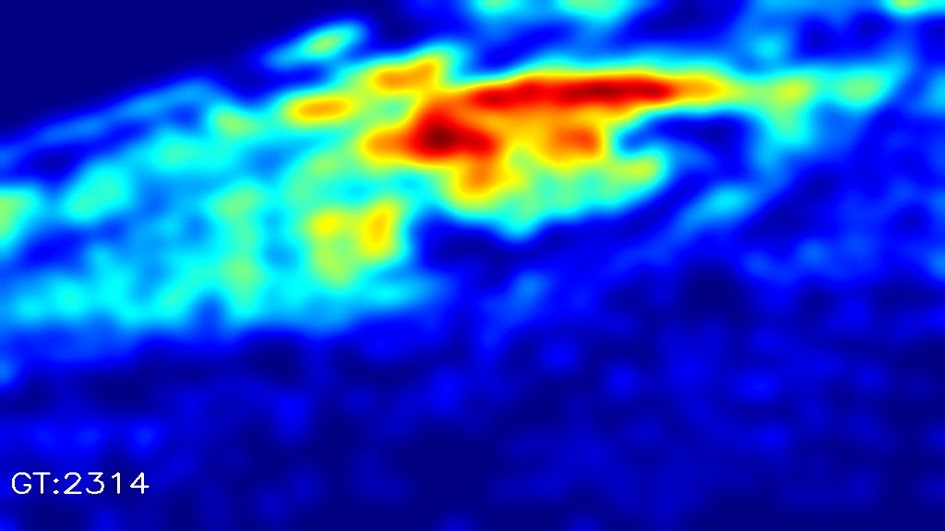}
    }
    \subfigure{
		\includegraphics[width=0.27\linewidth]{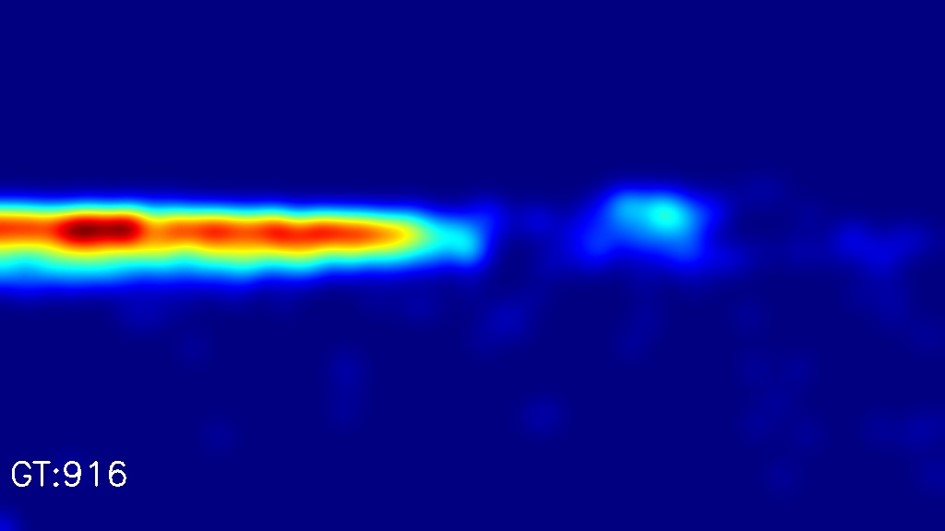}
    }
    \subfigure{
		\includegraphics[width=0.27\linewidth]{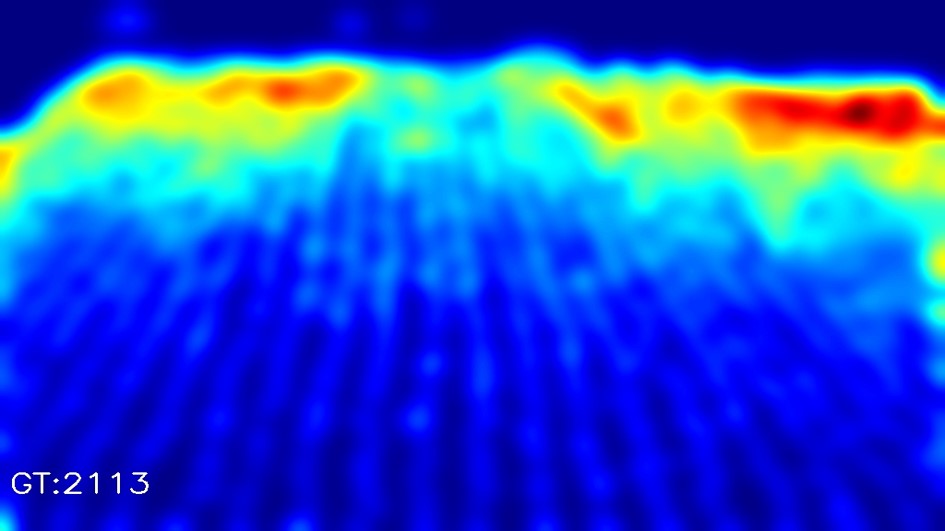}
    }
    \subfigure{
		\includegraphics[width=0.27\linewidth]{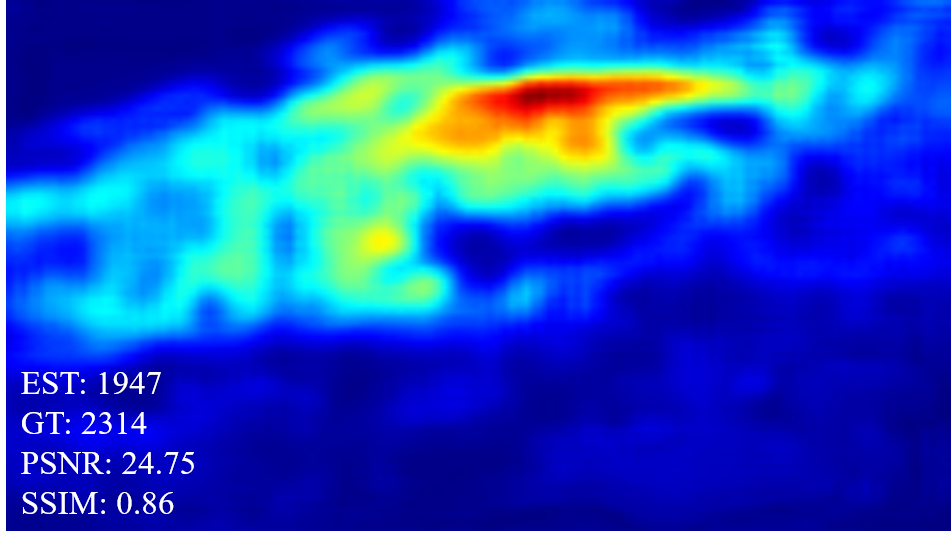}
    }
    \subfigure{
		\includegraphics[width=0.27\linewidth]{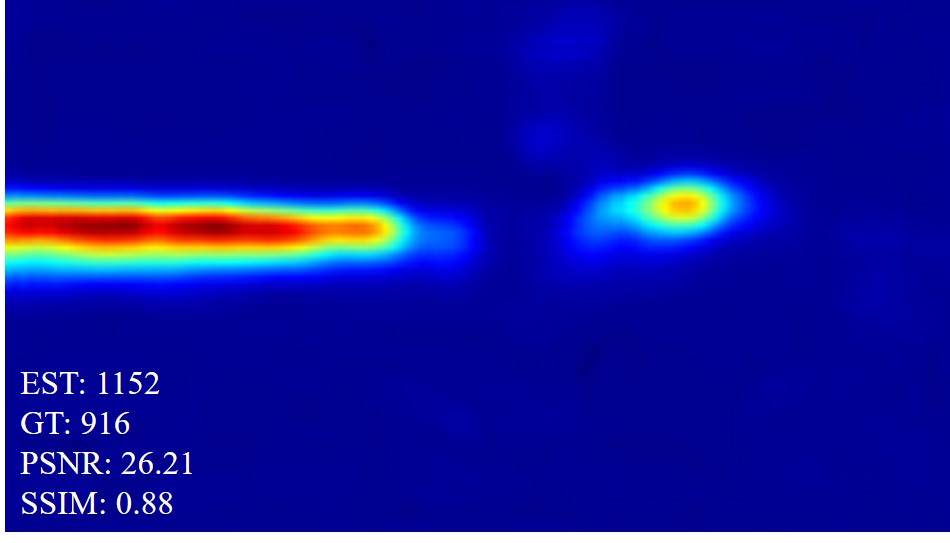}
    }
    \subfigure{
		\includegraphics[width=0.27\linewidth]{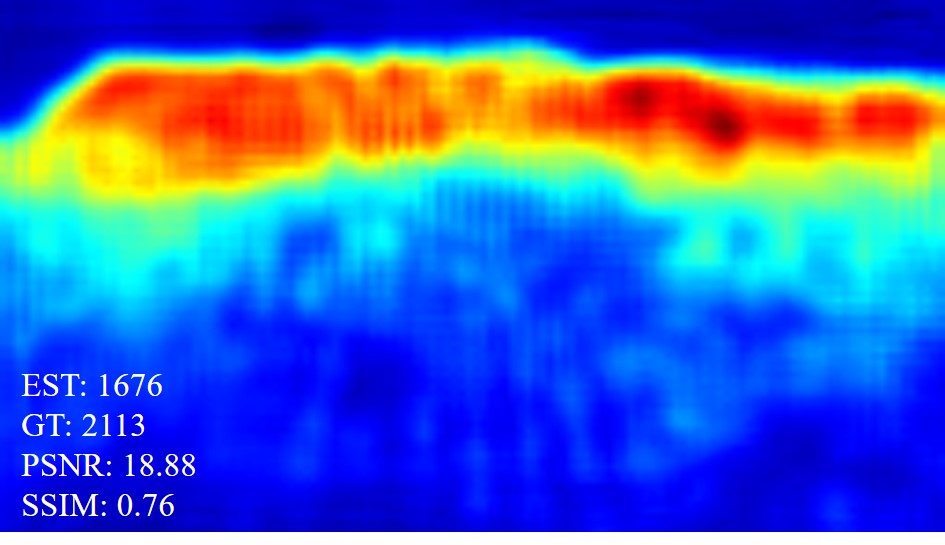}
    }
    \subfigure{
		\includegraphics[width=0.27\linewidth]{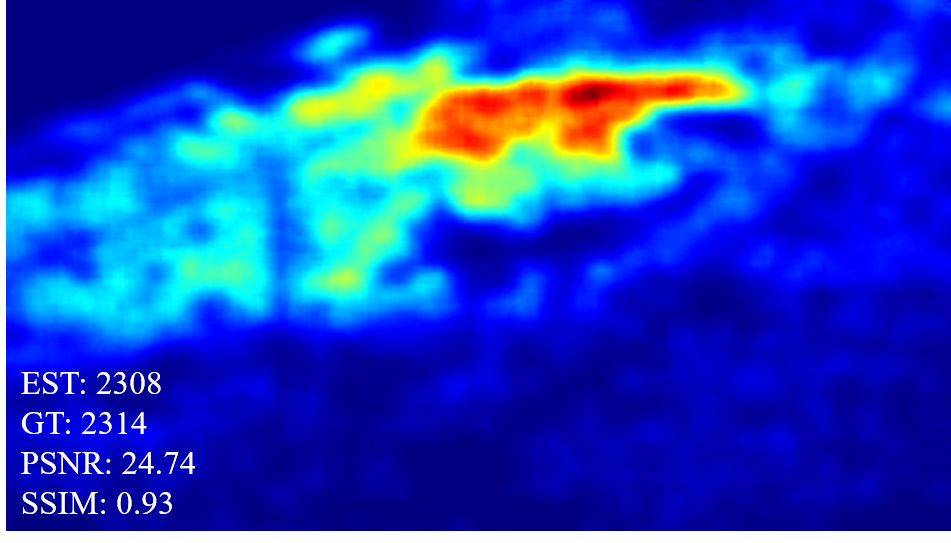}
    }
    \subfigure{
		\includegraphics[width=0.27\linewidth]{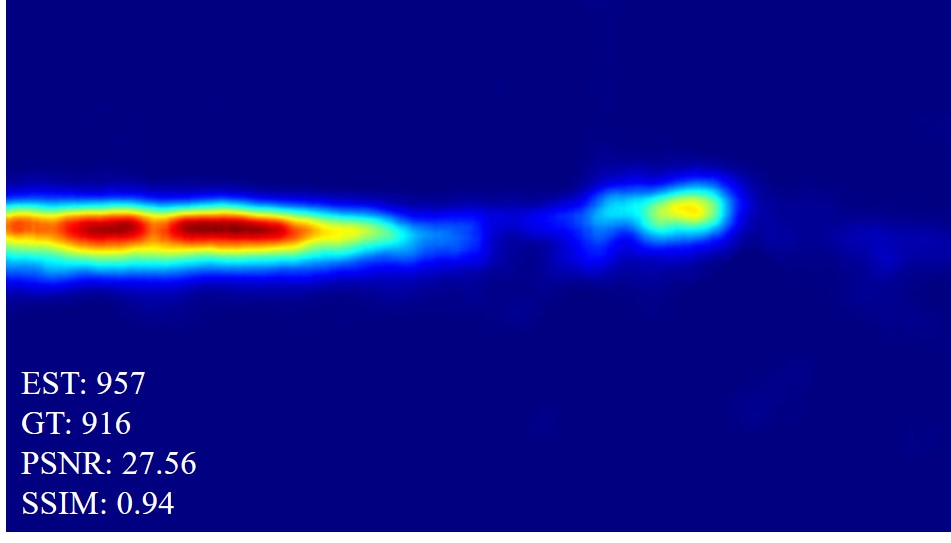}
    }
    \subfigure{
		\includegraphics[width=0.27\linewidth]{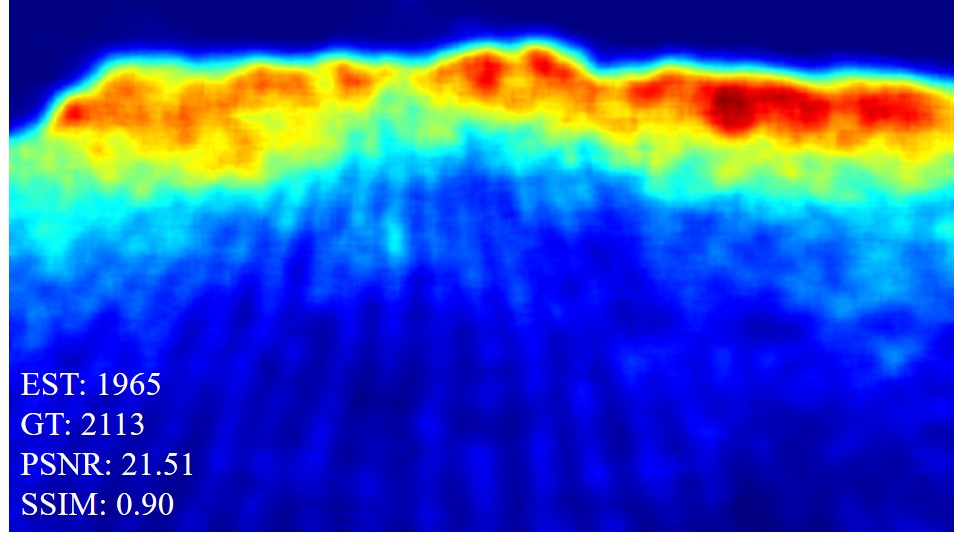}
    }
    \caption{Visualization of estimated density maps of UCF-QNRF Dataset. The first row presents the input image, the second row is the ground-truth density map, the third row is the predicted density map form CSRNet\cite{li2018csrnet}, and the fourth row shows the predicted density map of our method.}
    \label{vis_result}
\end{figure}

\subsection{Evaluations and Comparisons}
\subsubsection{ShanghaiTech Dataset}This dataset consists of 1198 images with a total of 330,165 annotation people, which is divided into Part A and Part B. The Part A includes 482 images with 300 images for training and 182 images for testing. The Part B contains 716 images and the number of training and testing images is 400 and 316, respectively.
As shown in Table \ref{SHT}, our method achieves an MAE of 60.1 and an RMSE of 100.2, which obtains the best performance among the listed methods. The result shows that the proposed method is able to aggregate multi-scale features for addressing  scale variations. 

\subsubsection{UCF\_CC\_50 Dataset} This dataset only contains 50 images of different sizes with 40 images for training and 10 images for testing, and the number of people in images ranges from 96 to 4633.
We perform the 5-fold cross-validation to evaluate our method. As shown in Table \ref{SHT}, the proposed method achieves an MAE of 181.3 and an RMSE of 258.6. Compared with SPN+L2SM\cite{xu2019learn}, our method improves 3.7\% in MAE and 17.9\% in RMSE. 

\subsubsection{UCF-QNRF Dataset}
The dataset is a large-scale crowd counting datatset with a total of 1.25 million annotated people head, which consists of 1535 high-quality images with 1201 images for training and 334 images for testing. 
Our proposed method obtains an MAE of 100.8 and an RMSE of 185.9, and it outperforms some state-of-the-art methods in table\ref{SHT} by a significant margin. This indicates that diverse contextual information is able to improve the counting accuracy of high-density area.

The visualization results of estimated density map on UCF-QNRF are showed in Figure \ref{vis_result}. Moreover, PSNR and SSIM metrics are used to compare the quality of predicted density map with other methods that focus on generating high-quality density map. As shown in Table \ref{SHHA-PSNR}, our method obtains a higher PSNR metric than other similar methods. This indicates the hierarchical attention-guided mechanism can improve the visual quality of density map.  
\begin{table}[t]
    \centering
    \caption{The evaluation of the quality of density map \\
    on ShanghaiTech Part A dataset.}
    \label{SHHA-PSNR}
    \begin{tabular}{|l|l|l|l|l|}
        \hline
        Methods     &att-map   &decoder  &PSNR$\uparrow$  &SSIM$\uparrow$ \\
        \hline
        Switch-CNN\cite{sam2017switching}  &$\times$  &$\times$  &21.91  &0.67 \\
        \hline
        CSRNet*\cite{li2018csrnet}     &$\times$  &$\times$  &23.04 &0.73 \\
        \hline
        MSPNet\cite{wei2020mspnet}      &$\surd$  &$\times$   &23.94 &0.78  \\
        \hline
        PCCNet\cite{gao2019pcc}      &$\surd$  &$\times$   &22.78 &0.74  \\
        \hline
        DADNet\cite{guo2019dadnet}      &$\surd$  &$\surd$   &24.16 &0.81  \\
        \hline
        ours        &$\surd$  &$\surd$   &25.04 &0.79  \\
        \hline        
    \end{tabular}
\end{table}

\subsection{Ablation Study}
In the ablation experiment, we use the first 10 layers of vgg16-bn network\cite{simonyan2014very} as our backbone to generate density map, which followed by two $3\times 3$ convolutional layers and one $1\times 1$ convolutional layer. 
Our baseline model obtains a strong benchmark result after introducing the batch normalization layers and data augmentation strategy. It achieves an MAE of 72.3 and an RMSE of 118, and outperforms most of early methods such as MCNN\cite{zhang2016single}, switch-CNN \cite{sam2017switching}. 
Besides, the backbone combined with the dense context-aware module (DCAM) can achieve an improvement of 13.3\% in MAE and 15.6\% in RMSE. This shows that the proposed DCAM can enhance the feature representation of model and handle the head scale variations better. 
Moreover, when we introduce a hierarchical attention-guided decoder, the quality of predicted density map can be significantly improved by 10.1\% in PSNR and 9.7\% in SSIM. This demonstrates that the detailed information from shallow feature maps can help to recover more spatial information of the density map.

\begin{table}[t]
    \centering
    \caption{Ablation study results on the ShanghaiTech Part A dataset.}
    \label{ABS}
    \begin{tabular}{|l|l|l|l|l|l|l|}
        \hline
        VGG   &DCAM      &HAG      &MAE$\downarrow$ &RMSE$\downarrow$ &PSNR$\uparrow$  &SSIM$\uparrow$ \\
        \hline      
        $\surd$    &$\times$  &$\times$  & 70.9       & 122.7   & 20.6      & 0.64 \\
        \hline
        $\surd$    &$\surd$   &$\times$  & 62.5      & 103.5    & 22.7        & 0.72 \\
        \hline
        $\surd$    &$\surd$   &$\surd$   & 60.1      & 100.2    & 25.0        & 0.79 \\
        \hline
    \end{tabular}
\end{table}

\section{Conclusion}
In this paper, we propose a novel crowd counting architecture called MSCANet, which helps to capture robust multi-scale features and generate a high-quality density map. The proposed DCAM densely connects multiple dilated kernels of different receptive fields to aggregate more diverse contextual information. To reduce background noise of density map,  the HAGD is designed to integrate different-level shallow features of the encoder through a multiple level supervision. Extensive experiments on three benchmarks show that the proposed method achieves competitive results than other state-of-the-art methods in terms of both counting accuracy and density map quality. 

\section*{Acknowledgment}
This work is supported partly by the projects of Guangdong Key Laboratory of Intelligent Information processing and Nation Key Research and Development Plan (2016YFB1200105, 2017YFB1302802, 2017YFB0902302).
\bibliographystyle{IEEEtran}
\bibliography{ref}

\end{document}